\documentclass[journal]{IEEEtran}
\usepackage{array}
\newcolumntype{P}[1]{>{\centering\arraybackslash}p{#1}}
\usepackage{cite}
\usepackage[pdftex]{graphicx}
\DeclareGraphicsExtensions{.pdf,.jpg,.png}
\usepackage{amsmath}
\usepackage{algorithmic}
\usepackage{graphicx}
\usepackage[numbers]{natbib}
\usepackage[justification=centering]{caption}
\usepackage{subcaption}
\usepackage{float}
\usepackage{amsmath}
\usepackage{color}

\usepackage{url}
\usepackage[none]{hyphenat}

\usepackage{fancyhdr}
\begin{document}
	\bstctlcite{IEEEexample:BSTcontrol}
	\title{Transfer Learning using CNN for Handwritten Devanagari Character Recognition}
	\author{
		\IEEEauthorblockN{
			Nagender Aneja\IEEEauthorrefmark{1} and Sandhya Aneja \IEEEauthorrefmark{1}
		} \\
		\IEEEauthorblockA{\IEEEauthorrefmark{1}Universiti Brunei Darussalam\\Brunei Darussalam
			\\\{nagender.aneja, sandhya.aneja
			\}@ubd.edu.bn}
	}
	
	\maketitle
	\thispagestyle{fancy}
	
	\begin{abstract}
		This paper presents an analysis of pre-trained models to recognize handwritten Devanagari alphabets using transfer learning for Deep Convolution Neural Network (DCNN). This research implements AlexNet, DenseNet, Vgg, and Inception ConvNet as a fixed feature extractor. We implemented 15 epochs for each of AlexNet, DenseNet 121, DenseNet 201, Vgg 11, Vgg 16, Vgg 19, and Inception V3. Results show that Inception V3 performs better in terms of accuracy achieving 99\% accuracy with average epoch time 16.3 minutes while AlexNet performs fastest with 2.2 minutes per epoch and achieving 98\% accuracy.
	\end{abstract}
	
	
	\IEEEpeerreviewmaketitle

	\section{Introduction}
\begin{figure}[!htbp]
		\centering
		\includegraphics[width=.9\linewidth]{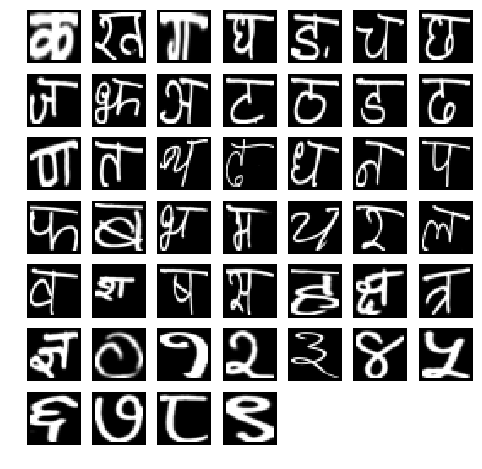}
		\caption{Devanagari Alphabets}
		\label{data}
	\end{figure}

	Handwriting identification is one of the challenging research domains in computer vision. Handwriting character identification systems are useful for bank signatures, postal code recognition, and bank cheques, etc. Many researchers working in this area tend to use the same database and compare different techniques that either perform better concerning the accuracy, time, complexity, etc. However, the handwritten recognition in non-English is difficult due to complex shapes. Many Indian Languages like Hindi, Sanskrit, Nepali, Marathi, Sindhi, and Konkani uses Devanagari script. The alphabets of Devanagari Handwritten Character Dataset (DHCD) \cite{acharya2015deep}  is shown in Figure \ref{data}.
	
\begin{figure}[!htbp]
		\centering
		\includegraphics[width=.5\linewidth]{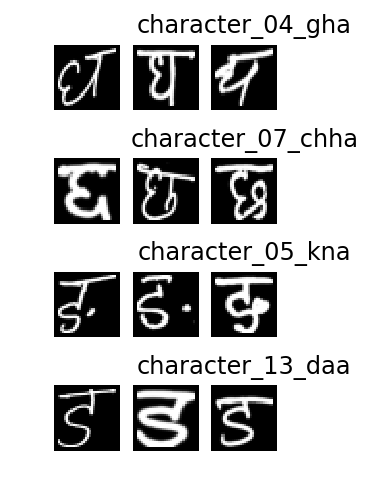}
		\caption{Similar Characters}
		\label{similar}
	\end{figure}

	Recognition of Devanagari is particularly challenging due to many groups of similar characters. For example, Figure \ref{similar} shows similarity is visible between character\_04 and character\_07; also between character\_05 and character\_13.
	
	Several researchers have proposed handcrafted features for recognition of strokes. Building a robust set of features for Devanagari is challenging due to the complex shape of characters. 
	
	In recent years, Deep Learning has gained popularity in computer vision and pattern recognition primarily due to Deep Convolution Neural Network (DCNN) \cite{anejaIoT}. 
	DCNN includes convolutional layers, fully connected layers, pooling layers for feature extraction, image representation, and dimensionality reduction respectively. DCNN also includes sigmoid or rectified linear units for non-linearity. CNN is based on local receptive fields and shared weights \citet{goodfellow2016deep}. The local receptive fields are regions that provide local information and are spread over the image. Each local receptive field constitutes a neuron and shares weights in the hidden layer. This helps to get distinct feature characteristics. Pooling layers that follow convolution layers, down samples and assist in overcoming overfitting by reducing the number of parameters.
	Additionally, dropout, wherein hidden neurons are randomly removed, also helps to avoid overfitting. Pre-trained CNN models are trained on \cite{imagenet} that carries a large number of data samples and 1000 class labels. The transfer learning techniques can help to apply the existing CNN models to problems where a large dataset is not available. 
	
	In this paper, we present Transfer learning, which is a deep learning technique that uses a pre-developed model for a task as a starting point for another task. Thus, it is a learning optimization technique that improves the performance of the second task. However, it only works if the model features learned from the first task are general. Transfer learning can be used either by (i) Develop Model Approach or (ii) Pre-trained Approach. In the develop model approach, the model is designed for one task and reused for the second task, however, in the pre-trained approach, one of the models available from different research organization is chosen as a starting point to tune it for the second problem. 
	
	This research compares various pre-trained models including \cite{alexnet, densenet, vgg, inception} in predicting Devanagari handwriting recognition. Following outlines are the main contributions of this paper:
	
	\begin{enumerate}
		\renewcommand{\theenumi}{\alph{enumi}}
		\item Inception model representations outperform in comparison to others regarding accurate results for our dataset.
		
		\item Vgg representations are also equally suitable for accurate results; however computationally expensive in comparison to Inception.  
		
		\item DenseNet representations did not show better accurate results but would be interesting to study over other datasets.
		
		\item AlexNet is fastest and also provides reasonably good accuracy
	\end{enumerate}
	
	\section{Related Work} \label{relatedwork}
	CNN based Deep Learning approaches have performed successfully in image processing; however, due to small data-size of Devanagari characters, the CNN models may overfit. The problem of training based on a small dataset has been solved by transfer learning.
	
	\citet{ng2015deep} classified human emotions in images extracted from movies. The authors implemented transfer learning from models AlexNet and Vgg and reported 48.5\% accuracy on the validation set.    
	
	\citet{chakraborty2018does} analyzed the accuracy of the deep neural network for handwritten Devanagari characters. The authors implemented CNN of five different depths. Two Bidirectional Long Short Term Memory (BLSTM) layers were added between the convolutional layers and the fully connected part of the five CNN networks. The authors reported the accuracy of 96.09\% with seven convolution layers and two fully connected layers.
	
	\citet{jangid2018handwritten} developed a deep convolutional neural network (DCNN) and adaptive gradient methods to identify handwritten Devanagari characters. The authors implemented six network architectures with different optimizers. The authors reported the accuracy of 96.02\% using NA-6 architecture and RMSProp optimizer.
	
	\citet{boufenar2018investigation} addressed Handwritten Arabic
	Character Recognition (HACR) using AlexNet \cite{krizhevsky2012imagenet} transfer learning approach in Matlab. The authors used a batch size of 100 images and 200 epochs and achieved 86.96\% accuracy. The authors obtained 99.33\% accuracy by fine-tuning the model.
	
	\citet{akccay2016transfer} used pre-trained AlexNet \cite{krizhevsky2012imagenet} and GoogLeNet \cite{szegedy2015going} to classify images for X-ray baggage security and achieved 98.92\% accuracy.

	\section{Experimental Setup}
	Handwriting recognition is an image classification problem, and this paper proposes the transfer learning and convolution neural network for this task \cite{oquab2014learning, yosinski2014transferable}. In particular, instead of designing a new deep convolutional neural network with the random initialization, we exploited existing trained DCNN on 1000 objects as a starting point to identify Devanagari alphabets.  We implemented AlexNet \cite{alexnet}; DenseNet 121 and DenseNet 201 \cite{densenet}; Vgg 11, Vgg 16, and Vgg 19 \cite{vgg}; and Inception V3 \cite{inception}. Training an entire DCNN with random initialization is highly computationally intensive and may not be practically feasible due to a limited dataset. Transfer Learning allows using the pretrained ConvNet either as initialization or fixed feature extractor. The pre-trained models are generally trained on ImageNet that has around 1.2 million images with 1000 different categories.
	
	Following are mainly two techniques that are used in transfer learning:
	
	\begin{enumerate}
		\item Finetuning: It includes instead of random initialization of weights of different layers, the network is initialized with a pretrained network, and the rest of the training takes place, as usual, using the new dataset
		
		\item Fixed Feature Extractor: In this approach, the weights of different convolutional layers are frozen that acts as feature extractor while the weights of the fully connected layer are updated. Thus in this approach only fully connected layer is trained.
	\end{enumerate}
	
	In some complex tasks, few upper convolutional layers are also trained in addition to fully connected layer. In this research, we used ConvNet as a fixed feature extractor.
	
	\subsection{Data Preparation}
	The dataset \cite{acharya2015deep} of Handwritten Devanagari characters has 46 classes with 2000 images of each class. We partitioned the dataset of 92,000 images into a training set of 78,200 images (0.85) and a testing set of 13,800 (0.15). The images have a resolution of 32 by 32 in the grayscale png format with the original character in the center. A padding of 2 pixels along sides was used in the images.
	
\begin{figure}[!htbp]
		\centering
		\includegraphics[width=.9\linewidth]{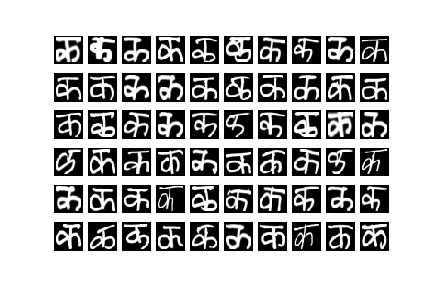}
		\caption{Sample images to demonstrate variation of Handwriting}
		\label{variation1}
	\end{figure}

	The dataset generated by \citet{acharya2015deep} had the handwriting of different people in wide variation that can be seen in Figure \ref{variation1}. The authors scanned handwritten documents and cropped each character manually to create this dataset of handwritten characters.
	
	We load the dataset into PyTorch tensors by resizing the images in size 224 by 224 and a normalizing using mean of (0.485, 0.456, 0.406) and standard deviation (0.229, 0.224, 0.225). We used PyTorch data generators to generate data in the batch size of 32.
	
	\subsection{Architectures}
	We used our base models on modern CNN architectures AlexNet, DenseNet, Vgg, and Inception. We first analyze the architecture of pre-trained models and their properties.
	
	\subsubsection{AlexNet} \citet{alexnet} has five convolution layers, three pooling layers, and two fully-connected layers. It was ILSVRC 2012 winner \cite{imagenet} with a DCNN that outperformed other handcrafted features resulting in a top-5 error rate of 16.4\%. This research extracted the features of convolution layers and trained the fully connected layers for Devanagari alphabets identification. 
	
	\subsubsection{DenseNet} With a large number of layers, gradient washes out; however, accuracy is better with a large number of layers.
	Each layer features are input to subsequent layers so to preserve the information extraction. CNN architecture proposed by \citet{densenet}  
	is interesting and useful for many datasets.
	
	\subsubsection{Vgg} \citet{vgg} showed that the small size convolutions with a preferred 3x3 filter size for large depth CNN networks  come up significantly accurate over CNN architectures proposed by  \citet{densenet} and  \citet{alexnet}.  Whereas a large number of parameters due to small convolutions incurs a high computational cost in comparison to prior-art configurations.
	
	\subsubsection{Inception} \citet{inception} extended the Vgg model by improving the filter blocks of size 3x3, 3x1,1x3 and 1x1 to reduce the number of parameters consequently a high reduction in computational cost.
	
	An overview of the comparison of the models used in this study is explained in Table \ref{comparison}.
	
	\begin{table*}[!htbp]
		\centering
		\caption{Comparison of Models}
		\label{comparison}
		\begin{tabular}{|P{2cm}|P{2cm}|P{2cm}|P{2.0cm}|P{2.0cm}|P{2.0cm}|}
			\hline
			Model  &  No of Parameters (in million)   & Top-1 Error  & Top-5 Error & Depth & No of crops  \\ \hline \hline
			AlexNet      & 60 &  37.5\% & 17.0\% &    8   & nil\\ \hline
			DenseNet 121 & 25 &  25.02\%      & 7.71\% &  100 &nil \\ \hline
			DenseNet 201 & 20 &  22.58\%      & 6.12\% &  190     & nil\\ \hline
			Vgg 11       & 134   &  24.7\%      & 7.5\% &    11   &1 \\ \hline
			Vgg 16       & 138   &    24.4\%     & 7.1\% &   16    &1 \\ \hline
			Vgg 19       & 144   &   23.7\%      & 6.8\% &   19    &1  \\ \hline
			Inception V3 & $\le$25& 21.2\% &  5.6\%  &100 & 1  \\ \hline
		\end{tabular}
	\end{table*}
	
	\subsection{Dataset and training}
	Training a CNN model with a limited amount of data without overfitting is a challenging task. However, the proposed approach of transfer learning using pre-trained CNN tackled the problem with significant improvement in performance. We trained the model using PyTorch by freezing the parameters of the convolutional layers of the pre-trained model while training the fully connected layer called classifier with the number of classes 46. We retained the original structure of the pre-trained model. Table \ref{hyperparameters} shows hyperparameters used in the training and machine configuration. 
	
	\begin{table*}[!htbp]
		\centering
		\caption{Parameters}
		\label{hyperparameters}
		\begin{tabular}{|P{3cm}|P{12cm}|}
			\hline
			{Parameter Name}  &  Values \\ \hline \hline
			Hyper Parameters &     Batch Size = 32, Learning rate = 0.001, Momentum = 0.9, Step size=7, Gamma=0.1, Epochs = 15 \\  \hline 
			Machine & i9 CPU, 64 GB RAM, Nvidia 1080 GPU, 1TB HDD \\ \hline
			Operating System & Ubuntu 18 \\ \hline
			Implementation & PyTorch and Torchvision \\ \hline
			Dataset & Training Files: 78200 and Testing Files: 13800 \\ \hline
			Image Size & 32x32 \\ \hline
			Image Type & PNG \\ \hline 
		\end{tabular}
	\end{table*}
	
	\section{Results and Discussion} \label{implementation}
	Table \ref{results3} summarizes the results. The results are explained regarding accuracy achieved in the first epoch, the best accuracy achieved in 15 epochs, epoch number in which we achieved the best accuracy, the total time taken by 15 epochs, average epoch time, and the number of features required by the model architecture.
	
	\begin{table*}[!htbp]
		\centering
		\caption{Comparison of Best Accuracy}
		\label{results3}
		\begin{tabular}{|P{1.8cm}|P{1.8cm}|P{1.8cm}|P{1.8cm}|P{1.8cm}|P{1.8cm}|P{1.8cm}|}
			\hline
			Model  &  Valid Accuracy (in 1st epoch) & Best Accuracy (in 15 epochs) & Best Accuracy acheiveed in \# epochs  & Total Time (15 epochs) & Average Training Time per Epoch & No of in features \\
			\hline \hline
			AlexNet & 95 & 98 &3& 33m 8sec & 2.2m &9216 \\ \hline
			DenseNet 121 & 73 & 89 &7 & 80 m 3 s & 5.3m &1024 \\ \hline
			DenseNet 201 & 74 & 90 &6 &113m 22s & 7.6m &1920 \\ \hline
			Vgg 11 & 97 & 99 &8& 86m 6s & 5.7m  &4096\\ \hline
			Vgg 16 & 97 & 98 &3& 132m 12s & 8.8m  &25088 \\ \hline
			Vgg 19 & 96 & 98&3 & 148m 57s & 9.9m   &25088 \\ \hline
			Inception V3 & 99 & 99&1 & 244m 36s & 16.3m  & 2048\\ \hline
		\end{tabular}
	\end{table*}
	
	Results show that Inception outperformed for our dataset with 99\% accuracy in the first epoch with the average time 16.3 minutes due to regularization imposed by highest number of layers and smaller convolution filter sizes. However, the computational cost of Inception is lower than second best model Vgg11 which shows 99\% accuracy in 45.6 minutes (time taken by eight epochs with average 5.7 minutes per epoch) as compared to one epoch of Inception in 16.3 minutes. The DenseNet model performed worst due to its architectural structure wherein the model feeds the connection between each layer to its subsequent layer. Since our dataset size is much smaller in comparison redundancy required for subsequent layers. AlexNet outperformed concerning computational cost with 98\% accuracy in 6.6 minutes with three epochs.

	\bibliographystyle{IEEEtranN}
	\bibliography{IEEEabrv,rpaper}

\end{document}